\documentclass{article}

\usepackage{microtype}
\usepackage{graphicx}
\usepackage{subfigure}
\usepackage{booktabs} 
\usepackage{hyperref}
\usepackage{bbm}

\usepackage[utf8]{inputenc} 
\usepackage[T1]{fontenc}    
\usepackage{hyperref}       
\usepackage{url}            
\usepackage{booktabs}       
\usepackage{amsfonts}       
\usepackage{nicefrac}       
\usepackage{microtype}      
\usepackage{natbib}

\usepackage{wrapfig}

\usepackage{hyperref}

\usepackage[preprint]{neurips_2020}
\usepackage{macros}

\usepackage{algorithmic}
\usepackage{algorithm}

\usepackage{xcite}
\usepackage{xr}
\makeatletter
\newcommand*{\addFileDependency}[1]{
  \typeout{(#1)}
  \@addtofilelist{#1}
  \IfFileExists{#1}{}{\typeout{No file #1.}}
}
\makeatother

\title{Partial Optimal Transport \\ with Applications on Positive-Unlabeled Learning}

\author{%
Laetitia Chapel\\
IRISA\\
Université Bretagne-Sud, IRISA, Vannes\\
\texttt{laetitia.chapel@irisa.fr}
\And
Mokhtar Z. Alaya\\
LITIS EA4108\\University Rouen Normandy\\
\texttt{mokhtarzahdi.alaya@gmail.com}
\And
Gilles Gasso \\
LITIS EA4108\\ University Rouen Normandy \& INSA Rouen  \\
\texttt{gilles.gasso@insa-rouen.fr}
}

\begin{document}
\maketitle




\begin{abstract}

Classical optimal transport  problem  seeks a transportation map that preserves the total mass betwenn two probability distributions, requiring their mass to be the same. This  may be too restrictive in certain applications such as color or shape matching, since the distributions may have arbitrary masses and/or that only a fraction of the total mass has to be transported. Several algorithms have been devised for computing partial Wasserstein metrics that rely on an entropic regularization, but when it comes with exact solutions, almost no partial formulation of neither  Wasserstein nor  Gromov-Wasserstein are available yet. This precludes from working with distributions that do not lie in the same metric space or when invariance to rotation or translation is needed. In this paper, we address the partial Wasserstein and Gromov-Wasserstein problems and propose exact algorithms to solve them. We showcase the new formulation in a positive-unlabeled (PU) learning application. To the best of our knowledge, this is the first application of optimal transport in this context and we first highlight that partial Wasserstein-based metrics prove effective in usual PU learning settings. We then demonstrate that partial Gromov-Wasserstein metrics is efficient in scenario where point clouds come from different domains or have different features.


\end{abstract}



\section{Introduction} 
\label{sec:introduction}

Optimal transport (OT)  has been gaining in recent years an increasing attention in the machine learning community. This success is due to its capacity to exploit the geometric property of the  samples at hand. Generally speaking, OT is a mathematical tool to compare distributions by computing a transportation mass plan from a source to a target distribution. Distances based on OT are referred to as the Monge-Kantorovich or Wasserstein distances
~\citep{villani03topics} and have been successfully employed in a wide variety of machine learning applications including clustering~\citep{ho2017}, computer vision~\citep{bonnel2011,solomon2015}, generative adversarial networks~\citep{pmlr-v70-arjovsky17a} or domain adaptation~\citep{courty2017optimal}.

\begin{figure}[htpb]
\includegraphics[width=1\textwidth]{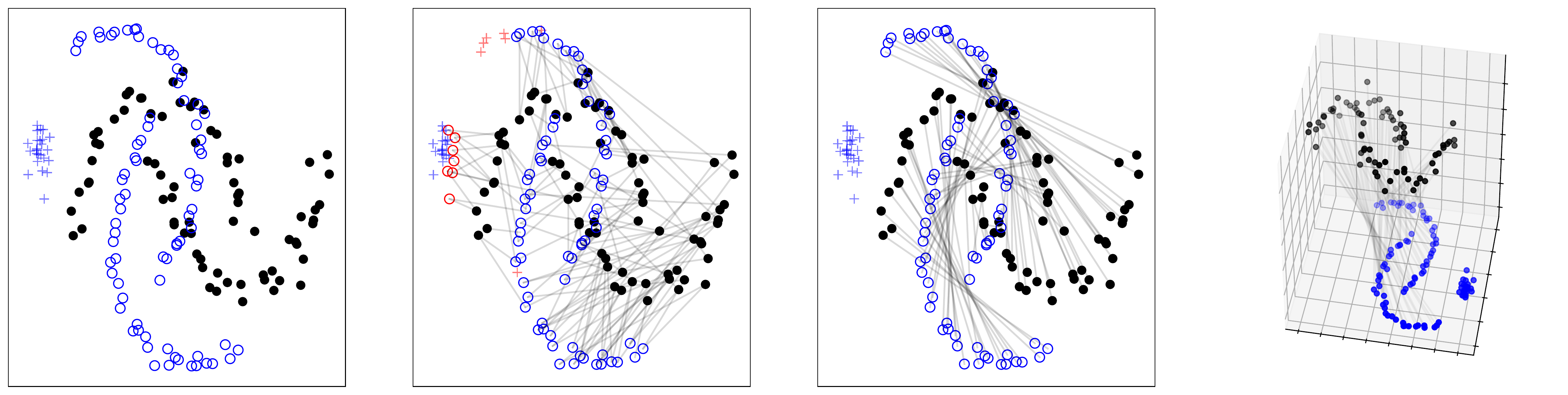}
\vskip -0.1in
\caption{(Left) Source (in black) and target (in blue) distributions that have been collected under distinct environments. The source domain contains only positive points ($o$) whereas the target domain contains both positives and negatives ($+$) (Middle left) Partial-W fails to assign correctly all the labels in such context, red symbols indicating wrong assignments (Middle right) Partial-GW recovers the correct labels of the unlabeled samples, with a consistent transportation plan (gray lines), even when the dataset do not live in the same state space (Right). }
\label{fig:poc2}

\end{figure}

A key limitation of Wasserstein distance is that it relies on the assumption of aligned distributions, namely they must belong to the same ground space or that at least a meaningful distance across domains can be computed. Nevertheless, source and target distributions can be collected under distinct environments, representing different times of collection, contexts or measurements (see Fig. \ref{fig:poc2}, left and right). To get benefit from OT on such heterogeneous distribution settings, one can compute the Gromov-Wasserstein (GW) distance~\citep{sturm2006,memoli2011GW} to overcome the lack of intrinsic correspondence between the distribution spaces.
GW extends Wasserstein by computing a distance between metrics defined within each of the source and target spaces. 
From a computational point view, it involves a non convex quadratic problem, hard to lift to large scale settings~\citep{peyre2019COTnowpublisher}. A remedy to a such heavy computation burden lies in a prevalent approach referred to as regularized OT~\citep{cuturinips13}, allowing one to add an entropic regularization penalty to the original problem. ~\citet{peyre2016Gromov-W} proposed the entropic GW discrepancy, that can be solved by Sinkhorn iterations~\citep{cuturinips13,benamou2015IterativeBP}.

A major bottleneck of OT in its traditional formulation is that it requires the two input measures to have the same total probability mass and/or that all the mass has to be transported. This is too restrictive for many applications since mass changes may occur due to a creation or an annihilation while computing an OT plan. To tackle this limitation, one may employ strategies such as {\it partial} or {\it unbalanced transport}~\citep{rubner2000,figalli2010,caffarelli2010}. ~\citet{chizat2018ScalingAF} propose to relax the marginal constraints of unbalanced total masses using divergences such as Kullback-Leibler or Total Variation, allowing the use of generalized Sinkhorn iterations. ~\citet{yang2018scalable} generalize this approach to GANs and ~\citet{lee.19c} present an ADMM algorithm for the relaxed partial OT. Most of all these approaches concentrate on partial-Wasserstein. 

This paper deals with exact partial Wassertein (partial-W) and Gromov-Wasserstein (partial-GW). Standard strategies for computing such partial-W require relaxations of the marginals constraints. We rather build our approach upon adding {\it virtual} or \textit{dummy} points onto the marginals, in the same spirit as \citet{caffarelli2010}. 
These points are used as a buffer when comparing distributions with different probability masses, allowing partial-W to boil down to solving an extended but standard Wasserstein problem. The main advantage of this approach is that it leads to computing sparse transport plans and hence exact partial-W or GW distances instead of regularized discrepancies obtained by running Sinkhorn algorithms.
Regarding partial-GW, our approach relies on a Frank-Wolfe optimization algorithm~\citep{frank-wolfe1956} that builds on computations of partial-W.

Tackling partial-OT problems that preserve sparsity
is motivated by the fact that they are more suitable to some applications such as the Positive-Unlabeled (PU) learning, see~\cite{bekker2018learning} for a review, we target in this paper. 
We shall notice that this is the first application of OT for solving PU learning tasks.
In a nutshell, PU classification is a variant of binary classification problem, in which we have only access to labeled samples from positive (\textbf{Pos}) class in the training stage. The aim is to assign classes to the points of an unlabeled (\textbf{Unl}) set which mixes data from both positive and negative classes. Using OT allows identifying the positive points within \textbf{Unl}, even when \textbf{Pos} and \textbf{Unl} samples do not lie in the same space (see Fig. \ref{fig:poc2}).

The paper is organized as follows: we first recall some background on OT. In Section 3, we propose an algorithm to solve an exact partial-W problem, together with a Frank-Wolfe based algorithm to compute the partial-GW  solution. After describing in more details the PU learning task and the use of partial-OT to solve it, we illustrate the advantage of partial-GW when the source and the target distributions are collected onto distinct environments. We finally give some perspectives. 

\paragraph{Notations}

$\Sigma_N$ is an histogram of $N$ bins with $\left\{\bp \in \mathbb{R}_+^N, \sum_i p_i = 1\right\}$  and $\delta$ is the Dirac function. Let $\mathbbm{1}_n$ be the $n$-dimensional vector of ones. $\inr{\cdot, \cdot}_F$ stands for the Frobenius dot product. $|\bp|$ indicates the length of vector $\bp$.

\section{Background on optimal transport} 
\label{sec:preliminaries}

Let  $\cX = \{\bx_i\}_{i=1}^{n}$ and $\cY=\{\by_j\}_{j=1}^{m}$ be two point clouds representing the source and target samples, respectively. We assume two empirical distributions $(\bp, \bq) \in \Sigma_n \times \Sigma_m$ over $\cX$ and $\cY$,   
\begin{equation*}
 \bp = \sum_{i=1}^{n} p_i\delta_{\bx_i} \quad \text{ and }\quad  \bq = \sum_{j=1}^{m} q_j\delta_{\by_j},
 \end{equation*} 
where $\Sigma_n$ and $\Sigma_m$ are histograms of $|\bp| = n$ and $|\bq| =m$ bins respectively. The set of all admissible couplings $\Pi(\bp,\bq)$ between histograms is given by
\begin{equation*}
	\Pi(\bp,\bq) = \{\bT \in \R_+^{|\bp|\times |\bq|} | \bT \mathbbm{1}_{|\bq|} = \bp, \bT^\top \mathbbm{1}_{|\bp|} = \bq\},
	\label{eq:admi_couplings}
\end{equation*}
where $\bT = \left(T_{ij}\right)_{i,j}$ is a coupling matrix with an entry $T_{ij}$ that describes the amount of mass $p_i$ found at $\bx_i$ flowing toward the mass $q_j$ of $\by_j$.

OT addresses the problem of optimally transporting $\bp$ toward $\bq$, given a cost $D_{ij}$ measured as a geometric distance between $\bx_i$ and $\by_j$. More precisely, when the ground cost $\bC= \mathbf{D}^p = \left(D_{ij}^p\right)_{i,j}$ is a distance matrix, the $p$-Wassertein distance on $\Sigma_n\times \Sigma_m$ at the power of $p$ is defined as:
\begin{equation*}
W_p^p(\bp, \bq) = \min_{\bT \in \Pi(\bp, \bq)}\inr{\bC, \bT}_F =  \min_{\bT \in \Pi(\bp, \bq)}\sum_{i=1}^n\sum_{j=1}^m C_{ij} T_{ij}.
\end{equation*}


\par
In some applications, the two distributions are not registered (\textit{i.e.} we can not compute a ground cost between $\bx_i$ and $\by_j$) or do not lie in the same underlying space. The Gromov-Wasserstein distance addresses this bottleneck by extending the Wasserstein distance to such settings, also allowing  invariance to translation, rotation or scaling. It relies on intra-domain distance matrices of source $\bC^s=(C^s_{ik})_{i,k} = (C^s(\bx_i, \bx_k))_{i,k}\in \mathbb{R}_+^{n\times n}$ and target $\bC^t= (C^t_{jl})_{j,l} = (C^t(\by_j, \by_l))_{j,l}\in \mathbb{R}_+^{m\times m}$, and is defined as in~\cite{memoli2011GW}:
\begin{equation*}
\label{GW}
GW_{p}^p(\bp, \bq) =\min_{\bT \in \Pi(\bp, \bq)} \sum_{i,k=1}^n\sum_{j, l=1}^m \left|C^s_{ik}-C^t_{jl}\right|^p  T_{ij}T_{kl}.
\end{equation*}

\section{Exact Partial Wasserstein and Gromov-Wasserstein distance} 
\label{sec:unbalanced_gromov_wasserstein}

\subsection{Partial Wasserstein distance}
\label{sec:partial_w}
The previous OT distances require the two distributions to have the same total probability mass  $\norm{\bp}_1 = \norm{\bq}_1$ and that all the mass has to be transported. This may be a problematic assumption where some mass variation or partial mass displacement should be handled. The partial OT problem focuses on transporting only a fraction $0 \leq s \leq \min(\norm{\bp}_1, \norm{\bq}_1)$ of the mass as cheaply as possible. In that case, the set of admissible couplings becomes
\begin{equation*}
	\Pi^u(\bp,\bq) = \{\bT \in \R_+^{|\bp|\times |\bq|} | \bT \mathbbm{1}_{|\bq|} \leq \bp, \bT^\top \mathbbm{1}_{|\bp|} \leq \bq, \mathbbm{1}_{|\bp|}^\top \bT \mathbbm{1}_{|\bq|}=s\},
	\label{eq:admi_couplings_partial}
\end{equation*}
and the partial-W distance reads as $PW^{p}_p(\bp, \bq) =  \min_{\bT \in \Pi^u(\bp, \bq)} \sum_{i=1}^n\sum_{j=1}^m \inr{\bC, \bT}_F.$
This problem has been studied by~\cite{caffarelli2010, figalli2010}; numerical solutions has notably been provided by~\cite{benamou2015IterativeBP, chizat2018ScalingAF} in the entropic-regularized Wasserstein case. 
We propose here to solve the exact partial-W problem by adding \textit{dummy} or \textit{virtual} points $\bx_{n+1}$ and $\by_{m+1}$ (with any features) and extending the cost matrix as follows:
\begin{equation}
\label{eq:dummypointC}
\bar{\bC} = \begin{bmatrix}
\bC & \xi \mathbbm{1}_{|\bq|}\\
 \xi \mathbbm{1}_{|\bp|}^\top & 2\xi+A
\end{bmatrix}
\end{equation}
in which $A > \max(C_{ij})$ and $\xi$ is a bounded scalar.
When the mass of these dummy points is set such that $p_{n+1} = \norm{\bq}_1 -s$ and $q_{m+1} = \norm{\bp}_1 -s$, computing partial-W distance boils down to solving a unconstrained problem $W_p^p(\bar{\bp}, \bar{\bq}) =  \min_{{\bar \bT} \in  \Pi(\bar{\bp}, \bar{\bq})}  \inr{\bar \bC, \bar \bT}_F$, 
where $\bar{\bp}=[\bp, \norm{\bq}_1 -s]$ and $\bar{\bq} =[\bq,  \norm{\bp}_1 -s]$. The intuitive derivation of this equivalent formulation is exposed in Appendix \ref{subsec:suppl_equivalence_PW_ExtW}.

\begin{proposition}
\label{approxpartial-extend-W}
Assume that $A > \max(C_{ij})$ and that $\xi$ is bounded, one has 
\begin{equation*}
W_p^p(\bar{\bp}, \bar{\bq}) - PW^{p}_p(\bp, \bq) = \xi (\norm{\bp}_1 + \norm{\bq}_1 - 2s)
\end{equation*}
and the optimum transport plan $\bT^*$ of the partial Wasserstein problem is the optimum transport plan $\bar{\bT}^*$ of $W_p^p(\bar{\bp}, \bar{\bq})$ deprived from its last row and column. 
\end{proposition}

The proof is postponed to Appendix~\ref{app:proof_propo_approx}. 

\subsection{Partial Gromov-Wasserstein}
We are now interested in the partial extension of Gromov-Wasserstein. In the case of a quadratic cost, $p=2$, and the partial-GW problem writes as 
\begin{equation*}
PGW_2^2(\bp, \bq) = \min_{\bT \in \Pi^{u}(\bp, \bq)}\cJ_{\bC^s, \bC^t}(\bT)
\end{equation*}
where 
\begin{equation}
\label{eq:JcsCt}
\cJ_{\bC^s, \bC^t}(\bT) = \frac 12 \sum_{i,k=1}^n\sum_{j, l=1}^m (C^s_{ik}-C^t_{jl})^2  T_{ij}T_{kl}.
\end{equation}
This loss function $\cJ_{\bC^s, \bC^t}$ is non-convex and the couplings feasibility domain  $\Pi^{u}(\bp, \bq)$ is convex and compact.
One may expect to introduce virtual points in the GW formulation in order to solve the partial-GW problem. Nevertheless, this strategy is no longer valid as it involves pairwise distances that do not allow the computations related to the dummy points to be isolated (see Appendix~\ref{appendix:partial_gw_dummpy_pts}).

In the following, we build upon a Frank-Wolfe optimization scheme~\citep{frank-wolfe1956} \textit{a.k.a.} the conditional gradient method~\citep{demyanovGC1973}. It has received significant renewed interest in machine learning~\citep{pmlr-v28-jaggi13,lacoste2015} and in OT community, since it serves as a basis to approximate penalized OT problems \citep{ferradans2013, courty2017optimal} or GW distances \citep{peyre2016Gromov-W,vayer2018fsw}. Our proposed Frank-Wolfe iterations strongly rely on computing partial-W distances and as such, achieve a sparse transport plan \citep{ferradans2013}.

Let us first introduce some additional notations. For any tensor $\cM = (\cM_{ijkl})_{i,j,k,l} \in \R^{n\times n\times m\times m}$, 
we denote by $\cM \circ \bT$  the matrix in $\R^{n\times m}$ such that its $(i,j)$-th element is defined as 
\begin{equation*}
(\cM \circ \bT)_{i,j} = \sum_{k=1}^{n}\sum_{l=1}^{m} \cM_{ijkl}T_{kl} 
\end{equation*}
for all $i=1, \ldots, n, j=1, \ldots, m.$
Introducing the $4$-th order tensor $\cM(\bC^s, \bC^t) = \frac 12 ((C^s_{ik} - C^t_{jl})^2)_{i,j,k,l},$ we notice that $\cJ_{\bC^s, \bC^t}(\bT) $, following \citet{peyre2016Gromov-W}, can be written as
\begin{align*}
\cJ_{\bC^s, \bC^t}(\bT) 
&= \inr{\cM(\bC^s, \bC^t)\circ \bT, \bT}_F.
\end{align*}
The Frank-Wolfe algorithm for partial-GW is shown in Algorithm~\ref{algo:gromov-wasserstein}. Like classical Frank-Wolfe procedure, it is summarized in three steps for each iteration $k$, as detailed below. A theoretical study of the convergence of the Frank-Wolfe based algorithm for partial-GW is given in Appendix~\ref{app:cvgFGW}. Also detailed derivation of the line search step is provided in the supplementary material. 
\paragraph{Step1} Compute a linear minimization oracle over the set $\Pi^{u}(\bp, \bq)$, \textit{i.e.},
\begin{align}
\label{algo:step-1-fw}
\widetilde{\bT}^{(k)} \gets \argmin_{\bT \in \Pi^{u}(\bp, \bq)} \inr{\nabla \cJ_{\bC^s, \bC^t}(\bT^{(k)}), \bT}_F,
\end{align}
To do so, we solve an extended Wasserstein problem with the ground metric $\nabla \cJ_{\bC^s, \bC^t}(\bT^{(k)})$ extended as in eq.~\eqref{eq:dummypointC}:
\begin{align}
\bar{\bT}^{(k)} \gets \argmin_{\bT \in \Pi(\bar{\bp}, \bar{\bq})} \inr{\bar{\nabla} \cJ_{\bC^s, \bC^t}(\bT^{(k)}), \bT}_F,
\end{align}
and get $\widetilde{\bT}^{(k)}$ from $\bar{\bT}^{(k)}$ by removing its last row and column. 
\paragraph{Step2} Determine optimal step-size $\gamma^{(k)}$ subject to 
\begin{equation}		
\label{algo:step-size-fw}
\gamma^{(k)} \gets \argmin_{\gamma \in [0, 1]} \{\cJ_{\bC^s, \bC^t}((1 - \gamma)\bT^{(k)} + \gamma\widetilde{\bT}^{(k)})\}.
\end{equation}	
It can be shown that $\gamma^{(k)}$ can take the following values, with $\bE^{(k)} = \widetilde{\bT}^{(k)} - {\bT}^{(k)}$:
\begin{itemize}
\item if $ \inr{\cM(\bC^s, \bC^t)\circ\bE^{(k)}, \bE^{(k)}}_F <0$ we have 
\begin{equation*}
\gamma^{(k)} = \left\{
    \begin{array}{ll}
        0 & \mbox{if } \inr{\cM(\bC^s, \bC^t)\circ\bE^{(k)}, \bE^{(k)}}_F+2\inr{\cM(\bC^s, \bC^t)\circ \bE^{(k)}, \bT^{(k)}}_F>0 \\
        1& \mbox{otherwise.}
    \end{array}
\right.
\end{equation*}
\item if $ \inr{\cM(\bC^s, \bC^t)\circ\bE^{(k)}, \bE^{(k)}}_F >0$ we have 
\begin{equation*}
\gamma^{(k)} = \min\left(1, \max\left(0, - \frac{  \inr{\cM(\bC^s, \bC^t)\circ \bE^{(k)}, \bT^{(k)}}_F }{\inr{\cM(\bC^s, \bC^t)\circ\bE^{(k)}, \bE^{(k)}}_F}\right)\right)
\end{equation*}
\end{itemize}
\paragraph{Step3} Update ${\bT}^{(k+1)} \gets (1- \gamma^{(k)}){\bT}^{(k)} + \gamma^{(k)}\widetilde{\bT}^{(k)}$.	

\begin{algorithm}[htb]
   \caption{Frank-Wolfe algorithm for partial-GW}
   \label{algo:gromov-wasserstein}
   \begin{algorithmic}[1]
    \STATE {\bfseries Input:} Source and target samples: $(\cX, \bp)$ and $(\cY, \bq)$, mass $s$, $p=2$, initial guess $\bT^{(0)}$\\
    \STATE Compute cost matrices $\bC^s$ and $\bC^t$,  build $\bar{\bp}$ and $\bar{\bq}$\\
	\FOR{$k=1,2,3, \ldots$}
	\STATE $\bG^{(k)} \gets \cM(\bC^s, \bC^t)\circ \bT^{(k-1)}$ \texttt{// Gradient computation}\\
	\STATE $\bar{\bT}^{(k)} \gets \argmin_{\bT \in \Pi( \bar{\bp}, \bar{\bq})} \inr{\bar{\bG}^{(k)}, \bT}_F$ \texttt{// Compute partial-W}\\
	\STATE Get $\widetilde{\bT}^{(k)}$ from $\bar{\bT}^{(k)}$ \texttt{// Remove last row and column}\\
	\STATE Compute $\gamma^{(k)}$ as in Eq. \eqref{algo:step-size-fw} \texttt{// Line-search}\\
	\STATE ${\bT}^{(k+1)} \gets (1- \gamma^{(k)}){\bT}^{(k)} + \gamma^{(k)}\widetilde{\bT}^{(k)}$ \texttt{// Update}\\
	\ENDFOR
    \STATE {\bfseries Return:} {${\bT}^{(k)}$}
\end{algorithmic}
\end{algorithm}


\section{Optimal transport for PU learning}
\label{sec:OT for PU}
We hereafter investigate the application of partial optimal transport for learning from Positive and Unlabeled (PU) data. After introducing the PU learning, we present how to adapt the formulation of partial-OT to this setting. 


\subsection{Overview of PU learning}

Learning from PU  data 
is a variant of classical binary classification problem, in which the training data consist of only positive  points, and the test data is composed of unlabeled positives and negatives. 
Let  $ {\bf{Pos}}=\{ \bx_i \}_{i=1}^{n_P}$ be the positive samples drawn according to the conditional distribution $p(\bx | y=1)$ and $ {\bf{Unl}}=\{ \bx_i^U \}_{i=1}^{n_U}$ the unlabeled set sampled according to the marginal $p(\bx) = \pi p(\bx | y=1) + (1-\pi) p(\bx | y=-1)$.   The true proportion of positives, called class prior, is $\pi = p(y=1)$ and $p(\bx | y=-1)$ is the distribution of negative samples which are all unlabeled.
The goal is to learn a binary classifier  solely using \textbf{Pos} and \textbf{Unl}. A broad overview of existing PU learning approaches can be seen in \citep{bekker2018learning}. \par

Most PU learning  methods commonly rely on the  selected completely at random (SCAR) assumption \citep{elkan2008learning}, which assumes 
that the labeled samples are drawn at random among the positive distribution, independently of their attributes. Nevertheless, this assumption is often violated in real-case scenarios and PU data are often subject to selection biases, \emph{e.g.} when part of the data may be easier to collect. Recently, a less restrictive assumption has been studied: the selected at random (SAR) setting \citep{bekker2018} which assumes that the positives are labeled according to a subset of features of the samples. \citet{kato2018learning} move a step further and consider that the sampling scheme of the positives is such that $p(o=1 | \bx, y=1)$ ($o=1$ means observed label) preserves the ordering over the samples induced by the posterior distribution $p(y=1 | \bx, y=1)$. Other approaches, as in  \citep{HsiehNS19}, consider a classical PU learning problem adjuncted with a small proportion of observed negative samples. Those negatives are selected with bias  following the distribution $p(\bx | y=-1)$.

\subsection{PU learning formulation using partial optimal transport}

We propose in this paper to use partial optimal transport to perform PU learning. In that context, the unlabeled points \textbf{Unl} represent the source distribution $\cX$ and the positive points \textbf{Pos} are the target dataset $\cY$. We set the total probability mass  to be transported $s = \pi$ as the proportion of positives in the unlabeled set.  
As such, the transport matrix $\bT$ should be such that the unlabeled positive points are mapped to the positive samples (as they have similar features or intra-domain distance matrices) while the negatives are discarded (in our context, they are not tranported at all).
We look for an optimal transport plan that belongs to the following set of couplings, assuming $n=n_U$, $m=n_P$, $p_i = \frac 1n$ and $q_j = \frac 1m$:
\begin{equation}
	\Pi^{PU}(\bp,\bq) = \{\bT \in \R_+^{|\bp|\times |\bq|} | \bT \mathbbm{1}_{|\bq|} =\{\bp, 0\}, \bT^\top \mathbbm{1}_{|\bp|} \leq \bq, \mathbbm{1}_{|\bp|}^\top \bT \mathbbm{1}_{|\bq|}=s\},
	\label{eq:admi_couplings_pu}
\end{equation}
in which $\bT \mathbbm{1}_{|\bq|} =\{\bp, 0\}$ means that $\sum_{j=1}^m T_{ij} = p_i$ exactly or 0, $\forall i$ to avoid matching a unlabeled negative with a positive. This set is not empty as long as $s\mod p_i =0$. Though the  positive samples \textbf{Pos} are assumed easy to label,  their features may be corrupted with noise or they may be mislabeled. Let assume $0 \leq \alpha \leq 1-s$, the noise level.
 

To solve the Wasserstein problem  with the admissible constraint set \eqref{eq:admi_couplings_pu} related to this PU learning, we adopt a regularized point of view of the partial-OT problem as in~\cite{courty2017optimal}:
\begin{equation}
\bar \bT^* = \argmin_{\bar{\bT} \in  \Pi(\bar{\bp}, \bar{\bq})} \sum_{i=1}^{n+1}\sum_{j=1}^{m+1} \bar{C}_{ij} \bar T_{ij} + \eta \Omega(\bar \bT)
\label{eq:equiv-ext-PU}
\end{equation}
where $p_i = \frac{1-\alpha}{n}$, $q_j = \frac{s+\alpha}{m}$, and $\bar{\bq}$, $\bar{C}_{ij}$ are defined as in Section \ref{sec:partial_w}, $\eta \geq 0$ is a regularization parameter and $\alpha$ is the percentage of \textbf{Pos} that we assume to be noisy (that is to say we do not want to match them with \textbf{Unl}). We choose  $\Omega(\bar \bT)=\sum_i\sum_{\text{g}}\norm{ \bar \bT (i, \mathcal{I}_{g})}_2$, where $\mathcal{I}_{g}$ contains the indices of the columns of $\bar \bT$ that correspond to either the positives ($g = [1,m]$) or the dummy point ($g = m+1$). 
This group-lasso regularization helps enforcing the constraint $\bT \mathbbm{1}_{|\bq|} =\{\bp, 0\}$. Also notice that the definition of $p_i$ and $q_j$ allows ensuring that only a probability mass of $s = \pi$ is moved towards \textbf{Pos}. When partial-GW is involved, we use this regularized-OT in the step $(i)$ of the Frank-Wolfe algorithm.
%

We can establish that solving problem \eqref{eq:equiv-ext-PU} provides the solution to PU learning using partial-OT.

\begin{proposition}
\label{approxpartial-PU-W}
Assume that $A > \max(C_{ij})$, $\xi$ is bounded, there exists a large $\eta>0$ such that: 
\begin{equation*}
W_p^p(\bar{\bp}, \bar{\bq}) - PUW^{p}_p(\bp, \bq) = \xi (1-s).
\end{equation*}
where $PUW^{p}_p(\bp, \bq) = \min_{\bT \in \Pi^{PU}(\bp,\bq)} \sum_{i=1}^n\sum_{j=1}^m C_{ij} T_{ij}$ and $W_p^p(\bar{\bp}, \bar{\bq}) = \min_{\bar\bT \in \Pi(\bar{\bp},\bar{\bq})} \sum_{i=1}^{n+1}\sum_{j=1}^{m+1} \bar{C}_{ij} \bar T_{ij}.$
\end{proposition}
The proof is postponed to Appendix~\ref{app:proof_propo2_approx}. 
%


\section{Experiments}
\label{sec:numerical_experiments}

\subsection{Experimental design}
We illustrate the behavior of partial-W and -GW on real datasets in a PU learning context. First, we consider a SCAR assumption, then a SAR one and finally a more general setting, in which the underlying distributions of the point clouds come from different domains, or do not belong to the same metric space. Algorithm \ref{algo:gromov-wasserstein} has been implemented using the Python Optimal Transport (POT) toolbox \citep{flamaryPOT}. 

Following previous works \citep{kato2018learning, HsiehNS19}, we assume that the class prior $\pi$ is known throughout the experiments;  
 otherwise, it can be estimated from $ \{ \bx_i \}_{i=1}^{n_P}$ and $ \{ \bx_i^U \}_{i=1}^{n_U}$ using off-the-shelf methods, \textit{e.g.} \citep{ramaswamy2016mixture, Plessis2017}.  
 For both partial-W and partial-GW, we choose $p=2$ and the cost matrices $\bC$ are computed using Euclidean distance. 
 
 We carry experiments on real-world datasets under the aforementioned scenarios. We rely on six datasets \texttt{Mushrooms}, \texttt{Shuttle}, \texttt{Pageblocks}, \texttt{USPS}, \texttt{Connect-4}, \texttt{Spambase} from the UCI repository\footnote{\href{https://archive.ics.uci.edu/ml/datasets.php}{\texttt{https://archive.ics.uci.edu/ml/datasets.php}}} (following \cite{kato2018learning}'s setting) and  \texttt{colored MNIST} \citep{arjovsky2019invariant} to illustrate our methods in SCAR and SAR settings respectively. We also consider the \texttt{Caltech office} dataset, which is a common application of domain adaptation \citep{courty2017optimal} to explore the effectiveness of our method on heterogeneous distribution settings. Additional illustration of the regularization term of equation~\eqref{eq:equiv-ext-PU} is given in Appendix~\ref{app:groupcontraints}.
 
 Whenever they contain several classes, these datasets are converted into binary classification problems  following \citet{kato2018learning}, and the positives are the samples that belong to the $y=1$ class.  For UCI and \texttt{colored MNIST} datasets, we randomly draw  $n_P = 400$ positive  and $n_U = 800$ unlabeled points among the remaining data. As the datasets are smaller, we choose $n_P = 100$ and $n_U = 100$ for \texttt{Caltech office} dataset. To ease the presentation, we only report results with class prior $\pi$ set as the true proportion of positive class in the dataset. We ran the experiments 10 times and report the mean accuracy rate (standard deviations are shown in Appendix \ref{sec:addfig}).  We test 2 levels of noise in \textbf{Pos}: $\alpha=0$ or $\alpha=0.025$, fix $\xi = 0$ and choose a large $\eta = 1e6$. 
 
 For the experiments, we consider unbiased PU learning method (denoted by \textsc{pu} in the sequel) \citep{du2014analysis} and the most recent and effective method to address PU learning with a selection bias (called \textsc{pusb} below)  that tries to weaken the SCAR assumption \citep{kato2018learning}.  Whenever possible (that is to say when source and target samples share the same features), we compare our approaches \textsc{p-w} and \textsc{p-gw} with \textsc{pu}  and \textsc{pusb}; if not, we are not aware of any competitive PU learning method able to handle different features in \textbf{Pos} and \textbf{Unl}. The GW formulation is a non convex problem and the quality of the solution is highly dependent on the initialization. We test several starting points for \textsc{p-gw} and report the result that gets the lowest loss (see Appendix~\ref{app:init} for the details).

\subsection{Partial-W and partial-GW in a PU learning under a SCAR assumption}
Under SCAR, the \textbf{Pos} dataset and the positives in \textbf{Unl} are assumed independently and identically drawn according to the distribution $p(\bx | y=1)$  from a set of positive points. 
 We experiment on the UCI datasets and Table \ref{tab:resultsscar} (top) summarizes our findings. Except for \texttt{Connect-4} and \texttt{spambase}, partial-W has similar results or consistently outperforms \textsc{pu} and \textsc{pusb} under a SCAR assumption. Including some noise has little impact on the results, exept for the \texttt{connect-4} dataset. 
 Partial-GW has competitive results, showing that relying on intra-domain matrices may allow  discriminating the classes. It nevertheless under-performs relatively to partial-W, as the distance matrix $\bC$ between \textbf{Pos} and \textbf{Unl} is more informative than only relying on intra-domain matrices.

\begin{table}[h]

\caption{Average accuracy rates on various datasets. \textsc{(g)-pw 0} indicates no noise and \textsc{(g)p-w 0.025} a noise of 0.025. Best values are indicated boldface. }
\vskip -0.1in
\label{tab:resultsscar}
\begin{center}
\begin{small}
\begin{sc}
\begin{tabular}{lccccccc}
\toprule
    dataset & $\pi$&    pu &  pusb &  p-W 0&  p-W 0.025&  p-GW 0&  p-GW 0.025 \\
\midrule
  mushrooms &  0.518 &  91.1 &  90.8 &        96.3 &           \textbf{ 96.4} &         95.0 &                93.1 \\
    shuttle &  0.786 &  90.8 &  90.3 &        \textbf{95.8} &            94.0 &         94.2 &                91.8 \\
 pageblocks &  0.898 &   92.1 &  90.9 &        \textbf{92.2} &            91.6 &         90.9 &                90.8 \\
       usps &  0.167 &  95.4&  95.1 &        \textbf{98.3} &            98.1 &         94.9 &                93.3 \\
  connect-4 &  0.658 &  \textbf{65.6} &   58.3 &        55.6 &            61.7 &         59.5 &                60.8 \\
   spambase &  0.394 &  \textbf{84.3} &  84.1 &        78.0 &            76.4 &         70.2 &                71.2 \\
\midrule
Original mnist &  0.1&97.9 & 97.8 &{\bf 98.8}  & 98.6& 98.2 & 97.9  \\
Colored mnist& 0.1 & 87.0  & 80.0 & 91.5 &91.5 & 97.3&\textbf{98.0} \\ 
\midrule
surf c$\rightarrow$surf c &0.1& 89.3 &89.4 & 90.0&\textbf{90.2} &87.2 &87.0 \\
surf c$\rightarrow$surf a &0.1& \textbf{87.7} & 85.6& 81.6& 81.8& {85.6}&{85.6} \\
surf c$\rightarrow$surf w&0.1&84.4 & 80.5& 82.2&82.0 &\textbf{85.6} & 85.0\\
surf c$\rightarrow$surf d&0.1& 82.0& 83.2&80.0 &80.0 &87.6 & \textbf{87.8}\\
decaf c$\rightarrow$decaf c &0.1&93.9& \textbf{94.8} & 94.0 &93.2 &86.4 &87.2 \\
decaf c$\rightarrow$decaf a&0.1& 80.5& 82.2 &80.2 & 80.2&\textbf{89.2} &88.8 \\
decaf c$\rightarrow$decaf w&0.1& 82.4&83.8 & 80.2& 80.2&\textbf{89.2} &88.6 \\
decaf c$\rightarrow$decaf d &0.1& 82.6& 83.6 & 80.8& 80.6& \textbf{94.2}& 93.2\\
\bottomrule
%
\end{tabular}
\end{sc}
\end{small}
\end{center}
\vskip -0.1in
\end{table}

\subsection{Experiments  under a SAR assumption}

The SAR assumption supposes that \textbf{Pos} is drawn according to some features of the samples. To implement such a setting, we inspire from  \citep{arjovsky2019invariant} and we construct a colored version of \texttt{MNIST}: each digit is colored, either in green or red, with a probability of $90\%$ to be colored in red.  The probability to label a digit $y=1$ as positive depends on its color, with only green $y=1$ composing the positive set. The \textbf{Unl} dataset is then mostly composed of red digits.
Results under this setting are provided in Table \ref{tab:resultsscar} (middle). When we consider a SCAR scenario, partial-W exhibits the best performance. However, its effectiveness highly drops when a covariate shift appears in the distributions $p(\bx | y=1)$ of the \textbf{Pos} and \textbf{Unl} datasets as in this SAR scenario. On the opposite, partial-GW allows maintaining a comparable level of accuracy as the discriminative information are preserved in intra-domain distance matrices.

\subsection{Partial-W and -GW in a PU learning with different domains and/or feature spaces}

To further validate the proposed method in a different context, we apply partial-W and partial-GW to a domain adaptation task. We consider the \texttt{Caltech Office} dataset, that consists of four domains: \texttt{Caltech 256} (\textsc{c}) \citep{griffin2007caltech}, \texttt{Amazon} (\textsc{a}), \texttt{Webcam} (\textsc{w}) and \texttt{DSLR} (\textsc{d})  \citep{saenko2010adapting}. There exists a high inter-domain variability as the objects may face different illumination, orientation \emph{etc}. 
Following a standard protocol, each image of each domain is described by a set of \textsc{surf} features \citep{saenko2010adapting} consisting of a normalized 800-bins histogram, and by a set of \textsc{decaf} features \citep{donahue2014decaf}, that are 4096-dimensional features extracted from a neural network.
The \textbf{Pos} dataset consists of images from \texttt{Caltech 256}. The unlabeled samples are formed by the  \texttt{Amazon}, \texttt{Webcam}, \texttt{DSLR} images together with the \texttt{Caltech 256} images that are not included in \textbf{Pos}. We perform a PCA to project the data onto $d=10$ dimensions for the \textsc{surf} features and $d=40$ for the \textsc{decaf} ones.

We first investigate the case where the objects are represented by the same features but belong to the same or different domains. Results are given in Table \ref{tab:resultsscar} (bottom). For both features, we first notice that \textsc{pu} and \textsc{pusb} have similar performances than partial-W when the domains are the same. As soon as the two domains  differ, partial-GW exhibits the best performances, suggesting that it is able to capture some domain shift.  We then consider a scenario where the source and target objects are described by different features (Table  \ref{tab:domainadaptation}). In that case, only partial-GW is applicable and its performances suggest that it is able to efficiently leverage on the discriminative information conveyed in the intra-domain similarity matrices, especially when using \textsc{surf} features to make predictions based on \textsc{decaf} ones. 
%

\begin{table}[!h]
\caption{Average accuracy rates on domain adaptation scenarios described by different features. As there is little difference between the results, we report performances only for $\alpha=0$.}
\label{tab:domainadaptation}
\vskip -0.1in
\begin{center}
\begin{small}
\begin{sc}
\begin{tabular}{lccccc}
\toprule
Scenario &  *=c & *=a & *=w& * = d\\
\midrule
surf c$\rightarrow$decaf * &87.0&94.4 & 94.4 & 97.4\\
decaf c$\rightarrow$surf * &85.0 &83.2 & 83.8 & 82.8 \\
\bottomrule
\end{tabular}
\end{sc}
\end{small}
\end{center}
\vskip -0.3in
\end{table}

%


\section{Conclusion and future work} 
\label{sec:conclusion}

The contributions of this work are twofold. 
We first show that considering an extended Wasserstein problem allows solving an exact partial-W. We then propose an algorithm relying on iterations of Frank-Wolfe method to solve a partial-GW problem, in which each iteration requires solving a partial-W problem. 
The second contribution relates to using partial-W and -GW distances to solve PU learning problems. We show that those distances compete and sometimes outperform the state-of-the-art PU learning methods, and that partial-GW allows remarkable improvements when the underlying spaces of the positive and unlabeled datasets are distinct or even unregistered. 
\par
While considering only features (with partial-W) or intra-domain distances (with partial-GW), this work can be  extended to define a partial-Fused Gromov-Wasserstein distance~\citep{vayer2018fsw} that can combines both aspects. Another line of work will also focus on lowering the computational complexity by using \textit{sliced} partial-GW, building on existing works on  \textit{sliced} partial-W~\citep{bonneel2019spot} and \textit{sliced} GW~\citep{vayer2019sliced}. Regarding the application view point, we envision a potential use of the approach to subgraph matching~\citep{Kriege2012subgraph} or PU learning on graphs \citep{zhao2011positive} as GW has been proved to be effective to compare structured data such as graphs. Finally, 
 we plan to derive an extension of this work to PU learning in which the proportion of positives in the dataset will be estimated in a unified optimal transport formulation, building on results of GW-based test of isomorphism between distributions~\cite{brecheteau2019statistical}.

\section*{Broader impact} 
\label{sec:broaderimpact}
This work does not present any significant societal, environnemental or ethical consequence. 

We have first defined a novel algorithm that gives the exact solution of the partial-(G)W problem. Several algorithms relying on a entropic regularization of the problem already exist and are known to be less computationally expensive than their exact counterpart. Nevertheless, in order to come close to a sparse solution, the regularization parameter should be very small, leading to a potential ill-conditioned optimal transport problem and a slow convergence of the algorithm. As such, the impact regarding the computational aspect is rather limited.

The second contribution is the use of OT to solve PU learning problems, even when the domains or features are not registered. This is of particular interest when the negative samples are difficult or costly to label and  PU learning allows in general the decrease of the labeling burden. Domain adaptation task may be considered as an interesting task to reduce the storage and labeling cost as data coming from different domains can be reused instead of collecting new and additional data.

\begin{ack}
This work was supported by grants from the MULTISCALE ANR-18-CE23-0022 Project and the OATMIL ANR-17-CE23-0012 Project of the French National Research Agency (ANR).
\end{ack}

\newpage
\bibliography{biblio}
\bibliographystyle{chicago}

\newpage
\appendix

\section{Partial Optimal Transport with dummy points}

The proof involves 3 steps:
\begin{enumerate}
\item we first justify the definition of $\bar\bp$ and $\bar\bq$ in the extended problem formulation, and show that $\bar{T}_{(n+1)(m+1)}$ should be equal to zero in order to have an equivalence between the original and the extended constraint set;
\item we then show that, for an optimal $\bar{\bT}^*$, we have $\bar{T}^*_{(n+1)(m+1)} = 0$ if $\bar{C}_{(n+1)(m+1)}>2\xi+ \max(C_{ij});$ 
\item we finally show that the solution of the extended Wasserstein problem $\bar{\bT}^*$ (deprived from its last row and column) and the one of the partial-Wasserstein one  ${\bT}^*$ are the same, and we show that $W_p^p(\bar{\bp}, \bar{\bq}) - PW_p^p(\bp, \bq) = \xi (\norm{\bp}_1 + \norm{\bq}_1 - 2s)$.
\end{enumerate}

\subsection{Equivalence between the constraint set of the extended Wasserstein problem and the partial-W problem} \label{subsec:suppl_equivalence_PW_ExtW}

Let recall the  formulation of partial OT problem that aims to transport only a fraction $0 \leq s \leq \min(\norm{\bp}_1, \norm{\bq}_1)$ of the mass as cheaply as possible. In that case, the  problem to solve is: 
\begin{equation*}
\min_{\bT \in \Pi^u(\bp, \bq)}\inr{\bC, \bT}_F 
\end{equation*}
with the constraint set:
\begin{equation}
	\Pi^u(\bp,\bq) = \big\{\bT \in \R_+^{|\bp|\times |\bq|} | \bT \mathbbm{1}_{|\bq|} \leq \bp, \bT^\top \mathbbm{1}_{|\bp|} \leq \bq, \mathbbm{1}_{|\bp|}^\top \bT \mathbbm{1}_{|\bq|}=s\big\}.
\label{eq:admi_couplings_partial}
\end{equation}

To express this as a standard discrete Kantorovitch optimal transport problem, we  get rid of the inequality constraints by re-writting (using standard tricks of linear programming):
\begin{eqnarray*}
\bT \mathbbm{1}_{|\bq|} + \bb & = &  \bp \\ 
\bT^\top \mathbbm{1}_{|\bp|} + \ba & = &  \bq\\ 
\end{eqnarray*} 
with $\bb\in \R_+^{|\bp|} $ and $\ba \in \R_+^{|\bq|} $ two unknown  vectors. Using the fact that $\mathbbm{1}_{|\bp|}^\top \bT \mathbbm{1}_{|\bq|}=s$, we get then the following equality constraint for marginal $\bq$:
\begin{eqnarray*}
	\mathbbm{1}_{|\bp|}^\top \bT \mathbbm{1}_{|\bq|}+ \mathbbm{1}_{|\bp|}^\top \bb  & = &  \mathbbm{1}_{|\bp|}^\top \bp \\
	s + \mathbbm{1}_{|\bp|}^\top \bb & = &  \|\bp\|_1 
\end{eqnarray*}
and for marginal $\bp$
\begin{eqnarray*}
	\mathbbm{1}_{|\bp|}^\top \bT \mathbbm{1}_{|\bq|} + \ba^\top\mathbbm{1}_{|\bq|} & = &  \bq^\top \mathbbm{1}_{|\bq|}  \\
	s + \ba^\top\mathbbm{1}_{|\bq|} & = &  \|\bq\|_1 
\end{eqnarray*}

The relations $s + \ba^\top\mathbbm{1}_{|\bq|}  =   \|\bp\|_1$ and $s + \mathbbm{1}_{|\bp|}^\top \bb  =   \|\bq\|_1 $ take into account the constraint related to the transported mass $s$ and we establish in subsection \ref{subsec_app:transport_mass} of the supplementary material that this mass is preserved whenever we solve the extended problem without the explicit constraint $\mathbbm{1}_{|\bp|}^\top \bT \mathbbm{1}_{|\bq|}=s$. 
Gathering these elements leads to this equivalent formulation
\begin{equation*}
	\min_{\bT \in \Pi^u(\bp,\bq)}   \inr{\bC, \bT}_F 
\end{equation*}
with $\Pi^u(\bp,\bq)$ re-expressed as
\begin{eqnarray*}
	\Pi^u(\bp,\bq) &= &\big\{\bT \in \R_+^{|\bp|\times |\bq|}~|~ \bT \mathbbm{1}_{|\bq|} +\bb = \bp, \bT^\top \mathbbm{1}_{|\bp|} +\ba =  \bq, \mathbbm{1}_{|\bp|}^\top \bT \mathbbm{1}_{|\bq|}=s,	\\
	&& \qquad \qquad \qquad \quad \ba \in \R_+^{|\bq|} , \ba^\top\mathbbm{1}_{|\bq|} =  \|\bq\|_1 - s,  \bb \in \R_+^{|\bp|},  \mathbbm{1}^\top_{|\bp|}\bb   =    \|\bp\|_1 - s\big\}.
\end{eqnarray*}
By defining the following augmented matrices and vectors
$$
\bar \bT = \begin{bmatrix}
\bT & \bb \\
\ba^\top & \beta
\end{bmatrix},  \quad \bar{\bp} = \begin{bmatrix}
\bp \\ \|\bq\|_1 - s
\end{bmatrix}, \quad \mbox{and} \quad 
\bar{\bq} = \begin{bmatrix}
\bq \\ \|\bp\|_1 - s
\end{bmatrix}
$$
we get this compact formulation
\begin{equation}
	\Pi^e(\bar \bp,\bar \bq) = \big\{\bar \bT \in \R_+^{|\bar \bp|\times | \bar \bq|} | \bar \bT \mathbbm{1}_{|\bar \bq|} = \bar \bp, \bar \bT^\top \mathbbm{1}_{|\bar \bp|} = \bar \bq, \beta = 0\big\}.
	\label{eq:extendedconstraintset}
\end{equation}
In the following, we show how solving a Wasserstein problem under the constraint set~\eqref{eq:extendedconstraintset} and how recovering the solution of the original partial problem (with the constraint set \ref{eq:admi_couplings_partial}) from it. 

\subsection{Proof of Proposition~\ref{approxpartial-extend-W}}
\label{app:proof_propo_approx}

Let us denote $\bar{T}^*$ the optimal coupling of the extended problem  
\begin{equation*}
\bar{\bT}^* = \argmin_{\bar{\bT} \in  \Pi(\bar{\bp}, \bar{\bq})} \sum_{i=1}^{n+1}\sum_{j=1}^{m+1} \bar{C}_{ij} {\bar T}_{ij}.
\end{equation*}
and recall that we set
\begin{equation*}
\bar{\bC} = \begin{bmatrix}
\bC & \xi\mathbf 1_m\\
\xi\mathbf 1_n^\top & 2\xi+A
\end{bmatrix}
\end{equation*}

\subsubsection{We first check that \texorpdfstring{$\mathbbm{1}_{n}^\top \bar{\bT}^{\backslash*} \mathbbm{1}_{m}=s$  when $\bar{T}^*_{(n+1)(m+1)} = 0$}{{1}_{n}^T {T}^{*} {1}_{m}=s$  when ${T}^*_{(n+1)(m+1)} = 0$}} 
\label{subsec_app:transport_mass}

Assume $\bar{\bT}^{\backslash*}$ is the matrix $\bar \bT^*$ with the last row and column removed.
Let us first suppose that $\bar{T}^*_{(n+1)(m+1)} = 0$. 
As a consequence and 
because of the constraints on the marginals, we have 
\begin{equation*}
\sum_{j=1}^m \bar{T}^*_{(n+1)j} =   \norm{\bq}_1 - s
\end{equation*}
 and 
 \begin{equation*}
 \sum_{i=1}^n \bar{T}^*_{i(m+1)} =   \norm{\bp}_1 - s.
 \end{equation*}
 We can also easily see that 
 \begin{equation*}
 \mathbbm{1}_{n+1}^\top \bar{\bT}^* \mathbbm{1}_{m+1}=\norm{\bp}_1+\norm{\bq}_1 - s
 \end{equation*}
  as $\norm{\bar{\bp}}_1 = \norm{\bar{\bq}}_1 = \norm{\bp}_1+\norm{\bq}_1 - s$. This implies that 
 \begin{align*} 
\mathbbm{1}_{n}^\top \bar{\bT}^{\backslash*} \mathbbm{1}_{m} &= \mathbbm{1}_{n+1}^\top \bar{\bT}^* \mathbbm{1}_{m+1} - \sum_{j=1}^m \bar{T}^*_{(n+1)j} - \sum_{i=1}^n \bar{T}^*_{i(m+1)} - \bar{T}^*_{(n+1)(m+1)} \\
  &=(\norm{\bp}_1+\norm{\bq}_1 - s) - (\norm{\bq}_1 - s) - (\norm{\bp}_1 - s) -   0\\
  &= s. 
 \end{align*}  

Hence, we have established that $\mathbbm{1}_{n}^\top\bar{\bT}^{\backslash*} \mathbbm{1}_{m} = s$.
 
 \subsubsection{Let show that \texorpdfstring{$\mathbbm{1}_{n}^\top \bar{\bT}^{\backslash*} \mathbbm{1}_{m}=s + \bar{T}^*_{(n+1)(m+1)}$  when $\bar{T}^*_{(n+1)(m+1)} \neq 0$}{{1}_{n}^T {T}^{*} {1}_{m}=sT^*_{(n+1)(m+1)}$  when $T^*_{(n+1)(m+1)} \neq 0$}} 
 \label{suppl:subsec-122}
Let us now suppose that $T^*_{(n+1)(m+1)} \neq 0$. This implies that
 \begin{align*} 
\mathbbm{1}_{n}^\top \bar{\bT}^{\backslash*} \mathbbm{1}_{m} &= \mathbbm{1}_{n+1}^\top \bar{\bT}^* \mathbbm{1}_{m+1} - \sum_{j=1}^{m+1} \bar{T}^*_{(n+1)j} - \sum_{i=1}^{n+1} \bar{T}^*_{i(m+1)} + \bar{T}^*_{(n+1)(m+1)} \\
  &=(\norm{\bp}_1+\norm{\bq}_1 - s) - (\norm{\bq}_1 - s) - (\norm{\bp}_1 - s) + \bar{T}^*_{(n+1)(m+1)} \\
  &= s +  \bar{T}^*_{(n+1)(m+1)}
 \end{align*}  
 
\subsubsection{We now show that \texorpdfstring{$\bar{T}^*_{(n+1)(m+1)} = 0 \text{ when  }\bar{C}_{(n+1)(m+1)}>2\xi+\max(C_{ij})$}{} }

We have 
\begin{align*}
\bar{\bT}^*&= \argmin_{\bar {\bT} \in  \Pi(\bar{\bp}, \bar{\bq})} \sum_{i=1}^{n+1}\sum_{j=1}^{m+1} \bar{C}_{ij} \bar {T}_{ij}\\
&=  \argmin_{\bar {\bT} \in  \Pi(\bar{\bp}, \bar{\bq})} \sum_{i=1}^{n}\sum_{j=1}^{m} \bar {T}_{ij} \bar{C}_{ij} +  \sum_{i=1}^{n} \bar {T}_{i(m+1)} \bar{C}_{i(m+1)} + \sum_{j=1}^{m} \bar {T}_{(n+1)j} \bar{C}_{(n+1)j}\\
&\qquad \qquad\quad + \bar {T}_{(n+1)(m+1)} \bar{C}_{(n+1)(m+1)} \\
&= \argmin_{\bar{\bT} \in  \Pi(\bar{\bp}, \bar{\bq})} \sum_{i=1}^{n}\sum_{j=1}^{m} {\bar T}_{ij} {C}_{ij} +   \xi\sum_{i=1}^{n} \bar {T}_{i(m+1)} +  \xi \sum_{j=1}^{m} \bar {T}_{(n+1)j} + \bar {T}_{(n+1)(m+1)} \bar{C}_{(n+1)(m+1)}\\ 
&= \argmin_{\bar{\bT} \in  \Pi(\bar{\bp}, \bar{\bq})} \sum_{i=1}^{n}\sum_{j=1}^{m} \bar {T}_{ij} {C}_{ij} +   \xi (\norm{\bp}_1-s - \bar {T}_{(n+1)(m+1)}) +  \xi(\norm{\bq}_1-s- \bar{T}_{(n+1)(m+1)})\\
&\qquad\qquad \quad +\bar {T}_{(n+1)(m+1)} \bar{C}_{(n+1)(m+1)}\\ 
&= \argmin_{\bar{\bT} \in  \Pi(\bar{\bp}, \bar{\bq})} \sum_{i=1}^{n}\sum_{j=1}^{m} \bar {T}_{ij} {C}_{ij} +   \xi (\norm{\bp}_1 + \norm{\bq}_1 - 2s)+(\bar{C}_{(n+1)(m+1)} -2\xi)\bar {T}_{(n+1)(m+1)}.
\end{align*}
Let us now suppose that $\bar{T}^*_{(n+1)(m+1)} >0$. It means that, $\forall \boldsymbol{\Gamma}$ that belongs to the admissible constraint set $\Pi(\bar{\bp}, \bar{\bq})$ such that $\boldsymbol{\Gamma}_{(n+1)(m+1)}=0$, we have:
\begin{align*}
\sum_{i=1}^{n}\sum_{j=1}^{m} \bar{T}^*_{ij} {C}_{ij} + &\xi (\norm{\bp}_1 + \norm{\bq}_1 - 2s)+(\bar{C}_{(n+1)(m+1)} -2\xi)  \bar{T}^*_{(n+1)(m+1)}\\
&\leq \sum_{i=1}^{n}\sum_{j=1}^{m} \boldsymbol{\Gamma}_{ij} {C}_{ij} +   \xi (\norm{\bp}_1 + \norm{\bq}_1 - 2s)
\end{align*}
that we can rewrite as
\begin{eqnarray*}
\sum_{i=1}^{n}\sum_{j=1}^{m} \bar{T}^*_{ij} {C}_{ij} +(\bar{C}_{(n+1)(m+1)} -2\xi)\bar{T}^*_{(n+1)(m+1)} &\leq& \sum_{i=1}^{n}\sum_{j=1}^{m} \boldsymbol{\Gamma}_{ij} {C}_{ij}. \text{ Then, }\\
(\bar{C}_{(n+1)(m+1)} -2\xi)\bar{T}^*_{(n+1)(m+1)} &\leq& \sum_{i=1}^{n}\sum_{j=1}^{m} \boldsymbol{\Gamma}_{ij} C_{ij} - \sum_{i=1}^{n}\sum_{j=1}^{m} T^*_{ij}{C}_{ij}
\end{eqnarray*}
When we have $\bar{C}_{(n+1)(m+1)}=2\xi+A$, with $A>\max(C_{ij})$, 
\begin{eqnarray*}
\bar{T}^*_{(n+1)(m+1)} &\leq& \dfrac{\max(C_{ij})\sum_{i=1}^{n}\sum_{j=1}^{m} \boldsymbol{\Gamma}_{ij}  - \sum_{i=1}^{n}\sum_{j=1}^{m} \bar{T}^*_{ij}{C}_{ij}}{(\bar{C}_{(n+1)(m+1)} -2\xi)}\\ 
\bar{T}^*_{(n+1)(m+1)} &\leq& \dfrac{\max(C_{ij})\sum_{i=1}^{n}\sum_{j=1}^{m} \boldsymbol{\Gamma}_{ij}  - \sum_{i=1}^{n}\sum_{j=1}^{m} \bar{T}^*_{ij}{C}_{ij}}{A}\\ 
\bar{T}^*_{(n+1)(m+1)} &\leq& \dfrac{\max(C_{ij})s  - \sum_{i=1}^{n}\sum_{j=1}^{m} \bar{T}^*_{ij}{C}_{ij}}{A}\\ 
\bar{T}^*_{(n+1)(m+1)} &\leq& \dfrac{\max(C_{ij})}{A}s  - \sum_{i=1}^{n}\sum_{j=1}^{m} \bar{T}^*_{ij}\dfrac{{C}_{ij}}{A}\\ 
\bar{T}^*_{(n+1)(m+1)} &\leq& \dfrac{\max(C_{ij})}{A}s  - \sum_{i=1}^{n}\sum_{j=1}^{m} \bar{T}^*_{ij}\\ 
\end{eqnarray*}
We have previously shown that $\sum_{i=1}^{n}\sum_{j=1}^{m} \bar{T}^*_{ij} = s +  \bar{T}^*_{(n+1)(m+1)}$. It implies that $\dfrac{\max(C_{ij})}{A}s < \sum_{i=1}^{n}\sum_{j=1}^{m} \bar{T}^*_{ij}$ as $\dfrac{\max(C_{ij})}{A} < 1$ and $ \sum_{i=1}^{n}\sum_{j=1}^{m} \bar{T}^*_{ij} = s$ by assumption. We then have
$\dfrac{\max(C_{ij})}{A}s - \sum_{i=1}^{n}\sum_{j=1}^{m} \bar{T}^*_{ij} <0$ and then 
\begin{equation*}
\bar{T}^*_{(n+1)(m+1)} \leq \dfrac{\max(C_{ij})}{A}s  - \sum_{i=1}^{n}\sum_{j=1}^{m} \bar{T}^*_{ij} <0
\end{equation*}
which contradicts the initial hypothesis $\bar{T}^*_{(n+1)(m+1)} >0$.\\
We can then conclude that $\bar{T}^*_{(n+1)(m+1)} =0$ when $\bar{C}_{(n+1)(m+1)}=2\xi+A$, with $A>\max(C_{ij})$.
%
 
 \subsubsection{We then prove Proposition~\ref{approxpartial-extend-W}}

Let us denote $\bar{\bT}^{\backslash*}$ the matrix $\bar{\bT}^*$ deprived from its last row and column. We then show that $\bar{\bT}^{\backslash*}= \bT^*$
where $\bT^*$ is the solution of the original partial-W problem:
\begin{equation*}
\bT^* = \argmin_{\bT \in \Pi^u(\bp, \bq)} \sum_{i=1}^n\sum_{j=1}^m C_{ij} T_{ij}.
\end{equation*}

We have 
\begin{eqnarray*}
\bar{\bT}^* &=& \argmin_{\bar {\bT} \in  \Pi(\bar{\bp}, \bar{\bq})} \sum_{i=1}^{n+1}\sum_{j=1}^{m+1} \bar{C}_{ij} \bar{T}_{ij}\\
&=& \argmin_{\bar {\bT} \in  \Pi(\bar{\bp}, \bar{\bq})} \sum_{i=1}^{n}\sum_{j=1}^{m}  {C}_{ij} \bar{T}_{ij}+   \xi (\norm{\bp}_1 + \norm{\bq}_1 - 2s)\\ 
&=&  \argmin_{\bar {\bT} \in  \Pi(\bar{\bp}, \bar{\bq})} \sum_{i=1}^{n}\sum_{j=1}^{m} {C}_{ij} \bar{T}_{ij} + \text{constant}\\
\end{eqnarray*}

As we have  $\bar{\bp}=[\bp, \norm{\bq}_1 -s]$ and $\bar{\bq} =[\bq,  \norm{\bp}_1 -s]$, we can write 
\begin{align*}
\sum_{i=1}^{n+1}{T}_{ij} = {q}_j, \forall j \leq m \implies  \sum_{i=1}^{n}{T}_{ij} \leq {q}_j \implies  ({\bar{\bT}}^{\backslash*})^\top \mathbbm{1}_{n} \leq \bq\\
 \sum_{j=1}^{m+1}{T}_{ij} = {p}_i ,\forall i \leq n \implies   \sum_{j=1}^{m}{T}_{ij} \leq {p}_i \implies  {\bar{\bT}}^{\backslash*} \mathbbm{1}_{m} \leq \bp
\end{align*}
We also have $ \sum_{i=1}^n\sum_{j=1}^m \bar{T}^{\backslash*}_{ij}=s$.
Finally, we can write that $\bar{\bT}^{\backslash*}$ belongs to the following constraint set:
\begin{equation*}
	\{\bar{\bT}^{\backslash*} \in \R_+^{n\times m} |\bar{\bT}^{\backslash*} \mathbbm{1}_{m} \leq \bp, (\bar{\bT}^{\backslash*})^\top \mathbbm{1}_{n} \leq \bq, \mathbbm{1}_{n}^\top \bar{\bT}^{\backslash*} \mathbbm{1}_{m}=s\}
\end{equation*}
which is the same like $\Pi^u(\bp, \bq)$. We then reach the result
\begin{equation*}
\bar{\bT}^{\backslash*} = \argmin_{{\bT} \in  \Pi^u({\bp}, {\bq})} \sum_{i=1}^{n}\sum_{j=1}^{m} {C}_{ij} {T}_{ij} = \bT^*
\end{equation*}

Finally we can write:
\begin{equation*}
W^p_p(\bar{\bp}, \bar{\bq}) - PW^p_p(\bp, \bq) = \xi (\norm{\bp}_1 + \norm{\bq}_1 - 2s).
\end{equation*}
as long as $\bar{C}_{(n+1)(m+1)}=2\xi+A$, with $A>\max(C_{ij})$ and
\begin{equation*}
W_p^p(\bar{\bp}, \bar{\bq}) = PW_p^p(\bp, \bq) 
\end{equation*}
when $\xi = 0$.
\subsubsection{We finally show how to construct \texorpdfstring{$\bar{\bT}^*$ from $\bT^*$}{${T}^*$ from $T^*$}}
We have 
\begin{equation*}
\bT^* = \argmin_{\bT \in \Pi^u(\bp, \bq)} \sum_{i=1}^n\sum_{j=1}^m C_{ij} T_{ij} = \argmin_{\bT \in \Pi^u(\bp, \bq)} \sum_{i=1}^n\sum_{j=1}^m \bar{C}_{ij} T_{ij} = \bar{\bT}^{\backslash*}
\end{equation*}
Setting $\bar{T}^*_{(n+1)(m+1)} = 0$, $\bar{T}^*_{i(m+1)} = \bar{p}_i - \sum_{j=1}^m\bar{T}^*_{ij}$ and $\bar{T}^*_{(n+1)j} = \bar{q}_j - \sum_{i=1}^n\bar{T}^*_{ij}$, we recover the constraint set $\Pi(\bar{\bp}, \bar{\bq})$ and finally $\bar{\bT}^*$.

\subsection{Adding dummy points to the GW problem does not solve a partial-GW problem}
\label{appendix:partial_gw_dummpy_pts}

While solving the partial-W problem can be achieved by adding dummy points and extending the cost matrix $\bC$ in an appropriate way, the same strategy can not be set up for solving GW. Indeed, the equivalence between the partial and the extended problem is permitted because we can ensure that (as long as $A > \max(C_{ij})$ and $\xi$ is a bounded scalar) $\bar{T}^*_{(n+1)(m+1)} = 0$, which implies that $\sum_{i=1}^n\sum_{j=1}^m \bar{T}^*_{ij} = s$. If we extend the intra-domain cost matrices $\bC^s$ and $\bC^t$ on the same pattern as follows
\begin{equation*}
\bar{\bC^s} = \begin{bmatrix}
\bC^s & \xi\mathbf 1_n\\
 \xi\mathbf 1_n^\top & A
\end{bmatrix}
\text{ and }
\bar{\bC^t} = \begin{bmatrix}
\bC^t & \xi\mathbf 1_m\\
 \xi\mathbf 1_m^\top & A
\end{bmatrix}
\end{equation*}
(with a large constant $A > 0$)
the GW formulation involves pairs of points.  We have
\begin{equation*}
\bar{\bT}^* = \argmin_{\bT \in  \Pi(\bar{\bp},\bar{\bq})} \sum_{i,k=1}^n\sum_{j, l=1}^m \left(C^t_{ik}-C^s_{jl}\right)^2  T_{ij}T_{kl} +  (\star)
\end{equation*}
where
\begin{eqnarray*}
 (\star) & =& 2\sum_{i,k=1}^n \Big(\left(C^t_{ik}-A\right)^2  T_{i(m+1)}T_{k(m+1)} + \sum_{j=1}^m \left(C^t_{ik}-\xi\right)^2  T_{ij}T_{k(m+1)}\Big)  \\
 & \quad + & 2\sum_{j,l=1}^m \Big( \left(A-C^s_{jl}\right)^2  T_{(n+1)j}T_{(n+1)l} + \sum_{i=1}^n \left(\xi - C^s_{jl}\right)  T_{ij}T_{(n+1)l}\Big)  \\
 & \quad + &2 \sum_{j=1}^m \left(A-\xi\right)^2  T_{(n+1)j}T_{(n+1)(m+1)}  +  2\sum_{i=1}^n  \left(\xi-A\right)^2  T_{i(m+1)}T_{(n+1)(m+1)}   \\
 & \quad + & 2 \sum_{i=1}^n  \sum_{j=1}^n \left(C^t_{ik}-\xi\right)^2  T_{ij}T_{(n+1)(m+1)}  + 0, 
\end{eqnarray*}
does not allows having $\bar{T}^*_{(n+1)(m+1)} = 0$, and hence $\sum_{i=1}^n\sum_{j=1}^m \bar{T}^*_{ij} \neq s$.

\section{Details of Frank-Wolfe algorithm for partial-GW}

\subsection{Line-search} 
\label{sub:line_search}

The step size in the line-search of Frank-Wolfe algorithm for partial-GW is given by 
\begin{equation*}
\gamma^{(k)}_{\min} \gets \argmin_{\gamma \in [0, 1]}\big\{\cJ_{\bC^s, \bC^t}\big((1 - \gamma)\bT^{(k)} + \gamma\widetilde{\bT}^{(k)}\big)\big\}.
\end{equation*}
Define $\bE^{(k)} = \widetilde{\bT}^{(k)} - {\bT}^{(k)}$ and the function $\phi:[0,1]\rightarrow \R$ such that  
\begin{align*}
\phi(\gamma^{(k)}) 
&= \cJ_{\bC^s, \bC^t}(\bT^{(k)}+ \gamma \bE^{(k)}).
\end{align*}
Straightforwardly, one has 
\begin{align*}
\phi(\gamma)&= \inr{\cM(\bC^s, \bC^t)\circ(\bT^{(k)}+ \gamma \bE^{(k)}), \bT^{(k)}+ \gamma \bE^{(k)}}_F\\
&= \inr{\cM(\bC^s, \bC^t)\circ\bT^{(k)}, \bT^{(k)}}_F +\gamma\inr{\cM(\bC^s, \bC^t)\circ\bT^{(k)}, \bE^{(k)}}_F+
\gamma\inr{\cM(\bC^s, \bC^t)\circ \bE^{(k)}, \bT^{(k)}}_F\\
&\quad + \gamma^2\inr{\cM(\bC^s, \bC^t)\circ\bE^{(k)}, \bE^{(k)}}_F.
\end{align*}
Since we choose a quadratic cost, $p=2$, then for any $\bT, \bR$ one has $\inr{\cM(\bC^s, \bC^t)\circ \bR, \bT}_F = \inr{\cM(\bC^s, \bC^t)\circ \bT, \bR}_F$ and we can then rewrite
\begin{align*}
\phi(\gamma) &= \gamma^2\inr{\cM(\bC^s, \bC^t)\circ\bE^{(k)}, \bE^{(k)}}_F+ 2\gamma \inr{\cM(\bC^s, \bC^t)\circ \bE^{(k)}, \bT^{(k)}}_F+ \inr{\cM(\bC^s, \bC^t)\circ\bT^{(k)}, \bT^{(k)}}_F.
\end{align*}
We then have to find $\gamma^{(k)}_{o}$ that minimises $\phi(\gamma) = a\gamma^2 + b\gamma +c$, with
\begin{equation*}
a = \inr{\cM(\bC^s, \bC^t)\circ\bE^{(k)}, \bE^{(k)}}_F, \quad b = 2\inr{\cM(\bC^s, \bC^t)\circ \bE^{(k)}, \bT^{(k)}}_F, \quad  c = \inr{\cM(\bC^s, \bC^t)\circ\bT^{(k)}, \bT^{(k)}}_F
\end{equation*}
with its derivative $\phi'(\gamma) = 2a\gamma + b$. 
This yields the following cases:
\paragraph{Case 1: $a < 0.$} In that case, $\phi(\gamma)$ is a concave function, whose minimum is reached either for $\gamma = 0$ or $\gamma =1$.
We have $\phi(0) = c >0$ and $\phi(1) = a+b+c$. We then have:
$$
\gamma^{(k)}_{\min} = \left\{
    \begin{array}{ll}
        0 & \mbox{if } a+b>0 \\
        1& \mbox{otherwise.}
    \end{array}
\right.
$$

\paragraph{Case 2: $a > 0.$} In that case, $\phi(\gamma)$ is a convex function, whose minimum on $[0,1]$ is reached for $\gamma^{(k)}_{\min} \in \{-\frac{b}{2a}, 0,1\}$.

\subsection{Convergence guarantee.}
\label{app:cvgFGW}
Intuitively a stationary point $\bT^o$ for partial-GW problem verifies that every direction in the polytope with origin $\bT^o$ is correlated with the gradient of the loss $\cJ_{\bC^s, \bC^t}(\cdot)$, namely
$\inr{\nabla \cJ_{\bC^s, \bC^t}(\bT^o); \bT - \bT^o}_F \geq 0$ for all $\bT \in \Pi^{u}(\bp, \bq).$
A good criterion to measure distance to a stationary point at iteration $k$ is the often used {\it Frank-Wolfe gap}, which is defined by 
\begin{equation*}
	g_{k} = \inr{\nabla \cJ_{\bC^s, \bC^t}(\bT^{(k)}), \bT^{(k)} - \widetilde{\bT}^{(k)}}_F.
\end{equation*}
Note that $g_{k}$ is always non-negative, and zero if and only if at a stationary point. Thanks to Theorem 1 in~\cite{lacoste2016} we have 
\begin{equation*}
	\min_{0 \leq k \leq K} g_k \leq \frac{\max(2 J_0, L\text{diam}(\Pi^{u}(\bp, \bq))^2)}{\sqrt{K+1}},
\end{equation*}
where $J_0 = \cJ_{\bC^s, \bC^t}(\bT^{(0)}) - \min_{\bT \in \Pi^{u}(\bp, \bq)}\cJ_{\bC^s, \bC^t}(\bT)$ defines the initial suboptimality, $L$ is a Lipschitz constant of $\nabla\cJ_{\bC^s, \bC^t}$ and $\text{diam}(\Pi^{u}(\bp, \bq))$ is the $\norm{\cdot}_F$-diameter of $\Pi^{u}(\bp, \bq)$ (see \eqref{diameter}). Therefore we can state the following lemma characterizing the convergence guarantee:
\begin{lemma}
\label{lem:FW_gap}
	The Frank-Wolfe gap $g_K = \min_{0 \leq k \leq K}g_k$ for the partial-GW loss $\cJ_{\bC^s, \bC^t}$  after $K$ iterations satisfies 
	\begin{equation}
	\label{upperbound_g_K}
		 g_K \leq \frac{2\max(J_0, \sqrt{2}s(\max(C^s_{ij}) + \max(C^t_{ij})))}{\sqrt{K+1}}
	\end{equation}
	where $s$ is the total mass to be transported and $\max(C^s_{ij})$ is the maximum value of cost matrix $\max(C^t_{ij})$, similarly for $\bC^t$.
\end{lemma}
Note that for the implementations, one can set $\max(C^s_{ij})=1=\max(C^t_{ij})$, hence the upper bound in~\eqref{upperbound_g_K} becomes more tight regarding a good initialization of Algorithm~\ref{algo:gromov-wasserstein}. This can be used to reduce significantly the initial suboptimality $J_0$. Furthermore, according to Theorem 1 in~\cite{lacoste2016}, Algorithm~\ref{algo:gromov-wasserstein} takes at most $\bigO(1 / \varepsilon^2)$ iterations to find an approximate stationary point with a gap smaller than $\varepsilon.$ 

{\it Proof of Lemma~\ref{lem:FW_gap}.}
Let us first calculate the diameter of the couplings set $\Pi^{u}(\bp, \bq)$ with respect to the Frobenieus norm $\norm{\cdot}_F$. One has 
\begin{align}
\label{diameter}
\text{diam}(\Pi^{u}(\bp, \bq)) = \sup_{(\bT, \bQ) \in \Pi^{u}(\bp, \bq)^2} \norm{\bT - \bQ}_F.
\end{align}
Using triangle inequality and the fact that $T_{ij}, Q_{ij}$ are probability masses that is $T_{ij}, Q_{ij} \in [0, 1]$, we get 
\begin{align*}
\norm{\bT - \bQ}_F^2 &\leq 2 \norm{\bT}_F^2 + 2 \norm{\bQ}_F^2\\
&\leq 2\sum_{i,j}^{n,m} T_{ij}^2 + 2\sum_{i,j}^{n,m} Q_{ij}^2\\
& \leq 2\sum_{i,j}^{n,m} T_{ij} + 2\sum_{i,j}^{n,m} Q_{ij}\\
& \leq 4s 
\end{align*}
where $s$ in the total mass to be transported. Thus $\text{diam}(\Pi^{u}(\bp, \bq)) \leq 2\sqrt{s}.$

For the Lipschitz constant of the gradient of $\cJ_{\bC^s, \bC^t}$ we proceed as follows: for any $\bT, \bQ \in \Pi^{u}(\bp, \bq)$ we have 
\begin{align*}
\norm{\nabla\cJ_{\bC^s, \bC^t}(\bT) - \nabla\cJ_{\bC^s, \bC^t}(\bQ)}_F^2
& = \sum_{i,j}^{n,m}\Big(\sum_{k,l}^{n,m}\cM_{ijkl}(T_{kl} - Q_{kl})\Big)^2\\
& \leq \sup_{i,j,k,l}\cM_{ijkl}^2 \sum_{i,j}^{n,m}\Big(\sum_{k,l}^{n,m}(T_{kl} - Q_{kl})\Big)^2\\
& \leq 4\sup_{i,j,k,l}\cM_{ijkl}^2 \norm{\bT-\bQ}_F^2.
\end{align*}
We have also 
\begin{align*}
\sup_{i,j,k,l}\cM_{ijkl}^2 = \frac 14 \sup_{i,j,k,l}(C_{ik}^s - C_{jl})^2
\leq \frac12 ((\max(C^s_{ij}))^2 + (\max(C^t_{kl}))^2).
\end{align*}
Hence the Lipschitz constant of $\nabla\cJ_{\bC^s, \bC^t}(\cdot)$ verifies $L \leq \sqrt{2}(\max(C^s_{ij}) + \max(C^t_{kl}))$. This gives the desired result.

\section{Equivalence between the regularized extended problem and the PU learning problem and proof of Proposition~\ref{approxpartial-PU-W}}
\label{app:proof_propo2_approx}
Let's denote $\bar{\bT}^*$ the optimal coupling of the extended problem  
\begin{equation}
\bar{\bT}^* = \argmin_{\bar{\bT} \in  \Pi(\bar{\bp}, \bar{\bq})} \sum_{i=1}^{n+1}\sum_{j=1}^{m+1} \bar{C}_{ij} \bar{T}_{ij} + \eta \Omega(\bar{\bT})
\label{supl:eq-ext-pu-pb}
\end{equation}
in which $\Omega(\bar{\bT})=\sum_i\sum_{\text{g}}\norm{ \bar{\bT} (i, \mathcal{I}_{\text{g}})}_2$, where $\mathcal{I}_{\text{g}}$ contains the indices of the columns of $\bar{\bT}$ that correspond to either the positives ($g=[1,\ldots, m]$) or dummy points ($g=m+1$).

\subsection{We first show that \texorpdfstring{$\sum_{i=1}^n\sum_{j=1}^m \bar{T}^{*}_{ij} = s$}{$\sum_{i=1}^n\sum_{j=1}^m T^{*}_{ij} = s$}} 
\label{suppl:subsec_3_1}

Let recall that by construction, we have $\bp^\top = \begin{bmatrix}
\bp, & \|\bq\|_1 -s 
\end{bmatrix}$ and $\bq^\top = \begin{bmatrix}
\bq, & \|\bp\|_1 -s 
\end{bmatrix}$. Also due to the definition of the marginals $p_i = \frac{1-\alpha}{n}$ for $i=1,\cdots, n$ and $q_j = \frac{s+\alpha}{m}$ for $j=1,\cdots, m$ we get $\norm{{\bp}}_1 =1 -\alpha$, $\norm{{\bq}}_1 =s + \alpha$.

Therefore we arrive at the results $\sum_{i=1}^{n+1}\sum_{j=1}^{m+1} \bar{T}^*_{ij} = 1$ by construction. Using the development in Section \ref{suppl:subsec-122} of the supplemental, we can establish that $\sum_{i=1}^{n+1} {\bar T}^*_{i(m+1)} = 1-\alpha - s$, and $\sum_{j=1}^{m+1} {\bar T}^*_{(n+1)j} = \alpha $. Thereon we attain $\sum_{i=1}^n\sum_{j=1}^m \bar{T}^*_{ij} = 1 - (1-\alpha - s + \alpha) = s$.

\subsection{Proof of Proposition~\ref{approxpartial-PU-W}} 

Given a solution $\bar{\bT}^*$ of the extended PU learning problem stated in Equation (\ref{supl:eq-ext-pu-pb}), we can write the objective function of the extended problem as
\begin{align*}
\sum_{i=1}^{n+1}\sum_{j=1}^{m+1} \bar{C}_{ij} \bar{T}^*_{ij} & =   \sum_{i=1}^{n}\sum_{j=1}^{m} {C}_{ij} \bar{T}^*_{ij} +  \xi\sum_{i=1}^{n} \bar{T}^*_{i(m+1)}  + \xi\sum_{j=1}^{m} \bar{T}^*_{(n+1)j} + (2\xi + A) \bar{T}^*_{(n+1)(m+1)}\\
& = \sum_{i=1}^{n}\sum_{j=1}^{m} {C}_{ij} \bar{T}^*_{ij} +  \xi (1-\alpha - s + \alpha) + 0\\
& = \sum_{i=1}^{n}\sum_{j=1}^{m} {C}_{ij} \bar{T}^*_{ij} +  \xi (1 - s).
\end{align*}

Let's then show that $(\bar{\bT}^*)_{i,j=1}^{n,m} = \bT^*$, where $\bT^*$ is the solution of the original PU problem:
\begin{equation*}
\bT^* = \argmin_{\bT \in \Pi^{PU}(\bp, \bq)} \sum_{i=1}^n\sum_{j=1}^m C_{ij} T_{ij}.
\end{equation*}

As shown before, we know that $\bar{T}^*_{(n+1)(m+1)} = 0$, hence we can rewrite the extended problem as 
\begin{align*}
\bar{\bT}^*  
&= \argmin_{\bar{\bT} \in  \Pi(\bar{\bp}, \bar{\bq})} \sum_{i=1}^{n} \sum_{j=1}^{m}{C}_{ij} \bar{T}_{ij} + \eta \sum_{i=1}^{n}\norm{\bar{T}_{i\bullet}}_2 + \eta \norm{\bar{T}_{(n+1)\bullet}}_2 + \eta \norm{\bar{T}_{\bullet(m+1)}}_1.
\end{align*}
We now turn this problem into its equivalent constrained form, namely there exists some $\lambda> 0$, related to the regularization parameter $\eta$, such that 
\begin{align*}
\bar{\bT}^*  = 
\begin{cases}
\argmin_{\bar{\bT} \in  \Pi(\bar{\bp}, \bar{\bq})}  \sum_{i=1}^{n} \sum_{j=1}^{m}{C}_{ij}\bar{T}_{ij}\\
\text{ s.t. } \sum_{i=1}^{n}\norm{\bar{T}_{i\bullet}}_2 + \norm{\bar{T}_{(n+1)\bullet}}_2 + \norm{\bar{T}_{\bullet(m+1)}}_1 \leq \lambda
\end{cases}
\end{align*}
Let $\bar{\bT}^{\backslash*} \in \R^{n\times m}$ such that $\bar{T}^{\backslash*}_{ij} = \bar{T}_{ij}$ for all $i,j \in \{1, \ldots, n\} \times \{1, \ldots, m\}.$
Since we have the polytope constraint $\sum_{j=1}^{m+1}\bar{T}_{ij} = p_i$, $\forall i$, this means that $\sum_{j=1}^{m}\bar{T}_{ij} \leq p_i$ and analagously $\sum_{i=1}^{n}\bar{T}_{ij} \leq q_j$.
Using the established results from section \ref{suppl:subsec_3_1} of the supplementary, we can derive that the matrix $\bar{\bT}^{\backslash*} \in \R^{n\times m}$ belongs to the following constraint set:
\begin{equation*}
	\Pi^u(\bp, \bq) = \{{\bQ} \in \R_+^{n\times m} |\bQ \mathbbm{1}_{m} \leq \bp, \bQ^\top \mathbbm{1}_{n} \leq \bq, \mathbbm{1}_{n}^\top \bQ \mathbbm{1}_{m}=s\}.
\end{equation*}
Let  define $\mathbf{u}_i = \bar{T}_{i\bullet} \in \R^{m}$ for all $i=1, \ldots, n$, $\mathbf{v}=\bar{T}_{(n+1)\bullet} \in \R^{m}$ and $\mathbf{w} = \bar{T}_{\bullet(m+1)} \in \R^{n}$. Then the problem can be re-formulated as
\begin{align*}
\bar{\bT}^*  = 
\begin{cases}
\argmin\limits_{\bar{\bT}^{\backslash*\in \Pi^u(\bp, \bq)}, \mathbf{u}_i, \mathbf{v}, \mathbf{w}}  \sum_{i=1}^{n} \sum_{j=1}^{m}{C}_{ij}\bar{T}^{\backslash*}_{ij}\\
\text{ s.t. } \sum_{i=1}^{n}\norm{\mathbf{u}_i}_2 + \norm{\mathbf{v}}_2 + \norm{\mathbf{w}}_1 \leq \lambda 
\end{cases}
\end{align*}
On the other hand, using the polytope constraints we have 
\begin{eqnarray*}
\sum_{i=1}^{n}\norm{\mathbf{u}_i}_2 = \sum_{i=1}^{n}\sqrt{\sum_{j=1}^{m}\bar{T}_{ij}^2} & \leq &  \sum_{i=1}^{n}\sqrt{\sum_{j=1}^{m}\bar{T}_{ij}} \\
 & \leq &  \sum_{i=1}^n\sqrt{p_i} = \sum_{i=1}^n\sqrt{\frac{1- \alpha}{n}} = \sqrt{n} \sqrt{1- \alpha}.
\end{eqnarray*}
Also we can derive an upper bound on the norm of $\mathbf{v}$
\begin{equation*}
\norm{\mathbf{v}}_2 = \sqrt{\sum_{j=1}^m \bar{T}_{(n+1)j}^2} \leq \sqrt{\sum_{j=1}^m \bar{T}_{(n+1)j}} = \sqrt{p_{n+1} } = \sqrt{\alpha},
\end{equation*}
and finally 
\begin{equation*}
\norm{\mathbf{w}}_1 = \sum_{i=1}^{n}\bar{T}_{i(m+1)} = q_{m+1} = 1 - \alpha -s.
\end{equation*}
Gathering those inequalities, we get
\begin{equation*}
\sum_{i=1}^{n}\norm{\mathbf{u}_i}_2 + \norm{\mathbf{v}}_2 + \norm{\mathbf{w}}_1 \leq \sqrt{n} \sqrt{1- \alpha} + \sqrt{\alpha} + (1 - \alpha -s).
\end{equation*}
Therefore, for any value of $\lambda$ such that $\lambda >  \lambda_b := \sqrt{n} \sqrt{1- \alpha} + \sqrt{\alpha} + (1 - \alpha -s)$, the group-lasso constraint
$\sum_{i=1}^{n}\norm{\mathbf{u}_i}_2 + \norm{\mathbf{v}}_2 + \norm{\mathbf{w}}_1 \leq \lambda$ is always satisfied for any choice of $\mathbf{u}_i, \mathbf{v}, \mathbf{w}$ verifying $\norm{\mathbf{v}}_1 = \alpha,$ $\norm{\mathbf{w}}_1 = 1-\alpha -s,$  and the marginal constraints
\begin{equation}
\label{conditionbis}
p_i =\norm{\mathbf{u}_i}_1 +\bar{T}_{i(m+1)}, \qquad \forall i.
\end{equation}
One can choose a particular sparse solution for $\mathbf{u}_i, \mathbf{v}$ as follows: 

\begin{itemize}
	\item if $\bar{T}_{i(m+1)}=0, \text{ for } i \in \{1, \ldots, n\}$, condition~\eqref{conditionbis} implies necessary that $\sum_{j=1}^m \bar{T}^{\backslash*}_{ij} = \sum_{j=1}^m \bar{T}_{ij}=\norm{\mathbf{u}_i}_1 =p_i$.
	\item If not, one can choose $\mathbf{u}_i$  such that $\norm{\mathbf{u}_i}_1 =0$, and hence $\sum_{j=1}^m \bar{T}_{ij}=\sum_{j=1}^m \bar{T}^{\backslash*}_{ij} =0.$
\end{itemize}
 
So it remains that  the solution of the constrained problem 
\begin{equation*}
\bar{\bT}^{\backslash*}  \in \Pi^{PU}(\bp, \bq) = \{{\bQ} \in \R_+^{n\times m} |\bQ \mathbbm{1}_{m} \leq \{\bp, 0\}, \bQ^\top \mathbbm{1}_{n} \leq \bq, \mathbbm{1}_{n}^\top \bQ \mathbbm{1}_{m}=s\},
\end{equation*}

that is   either $\sum_{j=1}^{m}\bar{T}_{ij}$ or $\bar{T}_{i(m+1)}$ is exactly 0. This means that there exists some value $\eta_b$, related to $\lambda_b= \sqrt{n} \sqrt{1- \alpha} + \sqrt{\alpha} + (1 - \alpha -s)$, such that for $\eta \geq \eta_b$, solving the extended problem (\ref{supl:eq-ext-pu-pb}) amounts to solving our PU learning formulation, which concludes the proof.



\section {Initialization of partial-W and -GW}
\label{app:init}
The partial-OT computation is based on a augmented problem with a dummy point and, as such, is convex. On the contrary, the GW problem is non-convex and, although the algorithm is proved to converge, there is no guarantee that the global optimum is reached. The quality of the solution is therefore highly dependent on the initialization. We propose to rely on an initial Wasserstein barycenter problem to build a first guess of the transport matrix. 

For partial-GW, as the $\bC^s$ and  $\bC^t$ matrices do not lie in the same ground space, we can not define a distance function between their members. Nevertheless, within a domain, we can build two sets of  ``homogeneous'' points. Instead of relying on a classical partitioning algorithm such as $k$-means, we propose to look for a barycenter with atoms $\bU_2=[\bu_2^1,\bu_2^1]$ and weights $\bb$ of the set $\bU$ that minimizes the following function:
\begin{equation}
\label{initialization}
f(\bb, \bU_2) =  W_p^p(\bb,\bq)
\end{equation}
over the feasible sets for $\bU_2$, where $\bb = [\pi, 1-\pi]$. In other word, we look for the set $\bU_2$ that allows having the most similar (in the Wasserstein sense) distribution $\bq$ than $\bU$. The induced transport matrix gives then two clusters, the one with mass $\pi$ serving as an initialization matrix for the GW problem.

Whenever possible (that is to say when \textbf{Pos} and \textbf{Unl} belong to the same space, we also initialize the GW algorithm with the solution of Partial-W, and the outer product of $\bar{\bp}$ and $\bar{\bq}$.
\section {Effect of the group constraints on the Wasserstein coupling.}
\label{app:groupcontraints}

\begin{figure}[!h]
  \begin{center}
    \includegraphics[width=0.7\textwidth]{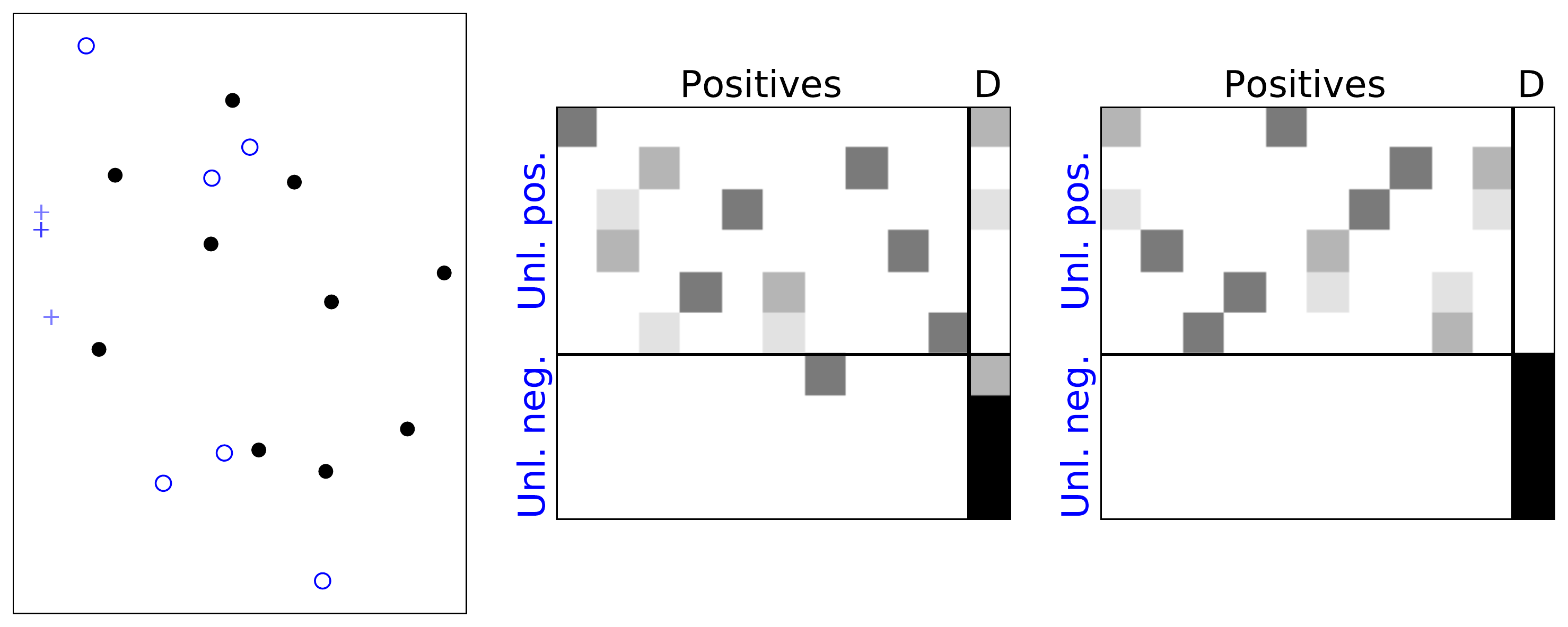}
  \end{center}
\label{fig:emdGroups}
\caption{(Left) Positives (black) and Unlabeled (blue) samples (Middle) Transport matrix with no group constraints, where darker color indicates stronger matching (Right) Transport matrix with group constraints.``D'' is the dummy point.}
  \vspace{-15pt}
\end{figure}

We first draw $n_P=10$  and $n_u=10$ samples, 6 of them being positives and we set $\alpha=0$. Fig. \ref{fig:emdGroups} shows the data and the obtained optimal couplings: enforcing a group constraint assigns some unlabeled points to the dummy point, allowing a clear identification of the negatives among \textbf{Unl}, whereas the solution with no such constraints may split the probability mass of the unlabeled positives between \textbf{Pos} and the dummy point, preventing to consistently identify the negatives. 

\section{Standard deviation of accuracy rates of the experiments}
\label{sec:addfig}
\begin{table}[!h]
\caption{Standard deviation of accuracy rates on different datasets and scenarios.}
\label{tab:domainadaptation_std}
\vskip 0.15in
\begin{center}
\begin{small}
\begin{sc}
\begin{tabular}{lccccccc}
\toprule
Dataset/Scenario &  $\pi$ & pu & pusb & p-w 0 &p-w 0.025 & p-gw 0 & p-w 0.025 \\
\midrule
Mushrooms  & 51.8 & 0.040& 0.047& 0.012 &             0.008 &          0.008 &              0.011 \\ %
Shuttle & 78.6 & 0.032 &0.045&0.009 &             0.012 &          0.018 &              0.017 \\ %
Pageblocks    & 89.8 &  0.012 & 0.010 &0.014 &             0.007 &          0.012 &              0.010\\ %
USPS    & 16.7 & 0.012 & 0.014&0.004 &             0.006 &          0.020 &              0.021 \\ %
Connect-4     & 65.8 & 0.003 & 0.010&0.017 &             0.016 &          0.019 &              0.017\\ %
Spambase     & 39.4 &  0.036&0.039&0.026 &             0.023 &          0.022 &              0.021 \\%
\bottomrule
Original mnist & 10 &0.005 & 0.005 &0.005  & 0.004 &0.004&0.01\\
Colored mnist& 10 & 0.035  & 0.034 & 0.004 &0.006&0.008&0.012\\ 
	\bottomrule	
surf c$\rightarrow$surf c & 10 &0.019& 0.015& 0.017 &0.014&0.015&0.017\\			
surf c$\rightarrow$surf a & 10 &0.014& 0.018& 0.012 &0.014&0.018&0.016\\
surf c$\rightarrow$surf w  & 10 &0.010& 0.064&0.014&0.013&0.016&0.010\\ 
surf c$\rightarrow$surf d     &10 &0.006 & 0.036&0.000&0.000&0.022&0.020\\
\midrule
decaf c$\rightarrow$decaf c &10 &0.022 & 0.017&0.013 &0.016&0.013&0.009\\
decaf c$\rightarrow$decaf a & 10 &0.019& 0.022& 0.006 &0.006&0.004&0.014\\
decaf c$\rightarrow$decaf w  &10 &0.030 & 0.009&0.006&0.006&0.014&\\ 
decaf c$\rightarrow$decaf d     & 10 &0.033& 0.008&0.010&0.009&0.015&0.003\\
\midrule

surf c$\rightarrow$decaf c &10 &- &- & - &-&0.012&0.013\\
surf c$\rightarrow$decaf a & 10 &-&- & - &-&0.011&0.011\\
surf c$\rightarrow$decaf w  &10 & -& -&-&-&0.006&006\\ 
surf c$\rightarrow$decaf d     &10 &- &- & - &-& 0.010&0.010\\

\midrule
decaf c$\rightarrow$surf c &10 & -& -& - &-&0.024&0.024\\
decaf c$\rightarrow$surf a     &10 &- &- & - &-&  0.037 &0.037\\
decaf c$\rightarrow$surf w    &10 &- &- & - &-&0.022&0.022\\
decaf c$\rightarrow$surf d     &10 &- &- & - &-&0.013&0.013\\
\bottomrule
\end{tabular}
\end{sc}
\end{small}
\end{center}
\vskip -0.1in
\end{table}

\end{document}